\documentclass[]{damo}

\usepackage{wrapfig}
\usepackage{tabularx}
\usepackage{textcomp}
\usepackage{stfloats}
\usepackage{url}
\usepackage{verbatim}
\usepackage{graphicx}
\usepackage{titlesec}
\usepackage{tocloft}
\usepackage{adjustbox}
\usepackage{multirow}
\usepackage{pifont}
\usepackage{tikz}
\usepackage{comment}
\usepackage{amsmath,amssymb} 
\usepackage{colortbl}  
\usepackage{color}
\usepackage{booktabs} 
\usepackage{hyperref}
\usepackage{graphicx}    
\usepackage{subcaption} 
\usepackage{booktabs}
\usepackage{amsmath} 
\usepackage{multirow} 
\usepackage{booktabs} 
\usepackage{subcaption} 
\RequirePackage{xspace}
\makeatletter
\DeclareRobustCommand\onedot{\futurelet\@let@token\@onedot}
\def\@onedot{\ifx\@let@token.\else.\null\fi\xspace}

\makeatother

\usepackage{makecell}

\definecolor{adptorange}{RGB}{248, 205, 172}
\definecolor{cmpblue}{RGB}{189, 215, 238}
\definecolor{cmpblue}{RGB}{189, 215, 238}

\definecolor{our_red}{RGB}{232,157,160}
\definecolor{our_blue}{RGB}{136,206,230}
\definecolor{our_orange}{RGB}{246,200,168}
\definecolor{our_green}{RGB}{178,211,164}

\definecolor{attn_code0}{RGB}{247,215,200}
\definecolor{attn_code1}{RGB}{238,169,139}
\definecolor{mlp_code0}{RGB}{204,201,221}
\definecolor{mlp_code1}{RGB}{102,95,153}

\definecolor{token_blue}{RGB}{84, 120, 140}

\definecolor{myMagenta}{rgb}{0.9,0,0.4}

\usepackage{pifont}       
\usepackage{bbding}       
\usepackage{fontawesome}
\usepackage{xspace}

\usepackage{float}

\newlength\savewidth

\newcolumntype{x}[1]{>{\centering\arraybackslash}p{#1pt}}
\newcolumntype{y}[1]{>{\raggedright\arraybackslash}p{#1pt}}
\newcolumntype{z}[1]{>{\raggedleft\arraybackslash}p{#1pt}}

\renewcommand{\paragraph}[1]{\vspace{1mm}\noindent\textbf{#1}}
\usepackage{colortbl}
\usepackage{xcolor}
\usepackage{wrapfig}

\renewcommand{\paragraph}[1]{\vspace{1.25mm}\noindent\textbf{#1}}

\usepackage{algorithm}
\usepackage{listings}

\definecolor{codeblue}{rgb}{0.25, 0.5, 0.5}
\definecolor{codekw}{rgb}{0.35, 0.35, 0.75}
\lstdefinestyle{Pytorch}{
    language = Python,
    backgroundcolor = \color{white},
    basicstyle = \fontsize{9pt}{8pt}\selectfont\ttfamily\bfseries,
    columns = fullflexible,
    aboveskip=1pt,
    belowskip=1pt,
    breaklines = true,
    captionpos = b,
    commentstyle = \color{codeblue},
    keywordstyle = \color{codekw},
}

\definecolor{green}{HTML}{009000}
\definecolor{red}{HTML}{ea4335}

\title{{\color{orange}D3}Grasp: {\color{blue}D}iverse and {\color{red}D}eformable {\color{green}D}exterous Grasping for General Objects}
\author[1,]{Keyu Wang}
\author[1]{Bingcong Lu}
\author[1, 2, \dagger]{Zhengxue Cheng}
\author[2]{Hengdi Zhang}
\author[1]{Li Song}

\affiliation[1]{Shanghai Jiao Tong University}
\affiliation[2]{Paxini Tech.}

\contribution[\dagger]{Corresponding author}

\abstract{
Achieving diverse and stable dexterous grasping for general and deformable objects remains a fundamental challenge in robotics, due to high-dimensional action spaces and uncertainty in perception. In this paper, we present \textbf{D3Grasp}, a multimodal perception-guided reinforcement learning framework designed to enable \textbf{D}iverse and \textbf{D}eformable \textbf{D}exterous \textbf{Grasp}ing. We firstly introduce a unified multimodal representation that integrates visual and tactile perception to robustly grasp common objects with diverse properties. Second, we propose an asymmetric reinforcement learning architecture that exploits privileged information during training while preserving deployment realism, enhancing both generalization and sample efficiency.
Third, we meticulously design a training strategy to synthesize contact-rich, penetration-free, and kinematically feasible grasps with enhanced adaptability to deformable and contact-sensitive objects.
Extensive evaluations confirm that D3Grasp delivers highly robust performance across large-scale and diverse object categories, and substantially advances the state of the art in dexterous grasping for deformable and compliant objects, even under perceptual uncertainty and real-world disturbances. D3Grasp achieves an average success rate of \textbf{95.1}\% in real-world trials—outperforming prior methods on both rigid and deformable objects benchmarks.

}

\date{\today} 
\homepage{https://readerek.github.io/D3.github.io/}

\begin{document}
\thispagestyle{firstheader}
\maketitle
\pagestyle{empty}

\section{Introduction} \label{sec:introduction}
Dexterous robotic hands, with their human-like kinematic structures and multi-finger adaptability, hold transformative potential across industrial assembly, elderly care, and hazardous material handling. Recent advances in hardware design, exemplified by Shadow Dexterous Hand \cite{shadowhand2013}, Allegro Hand \cite{allegrohand}, and Paxini Dexhand13 \cite{paxinidexhand}, have enabled 16+ degree-of-freedom (DoF) manipulation capabilities approaching human-level dexterity. However, two fundamental challenges persist in bridging this mechanical potential to real-world applications \cite{xiao2025robot,an2025dexterous}: multimodal perception integration and data-efficient policy learning, particularly for long-horizon manipulation tasks.
Contemporary robotic manipulation systems primarily depend on single sensing modality, each with inherent limitations: vision enables global localization but struggles with transparency or occlusion; tactile sensing offers precise contact feedback, yet lacks global awareness; proprioception monitors internal states but provides minimal environmental understanding. Hybrid architectures, such as visual-tactile fusion networks ~\cite{vtli2024comprehensive,vtakinola2024tacsl,vtdave2024multimodal,vtferrandis2024learning,vtjin2023vision,vtparsons2022visuo}, attempt to address these constraints through direct sensor concatenation. However, this approach induces a high-dimensional observation space, hindering policy convergence \cite{tao2024curriculum}. Crucially, fixed fusion weights cannot adapt to the varying sensory dominance across manipulation phases \cite{li2022see,wang2025chain}, often resulting in conflicting signals that degrade control stability \cite{vtakinola2024tacsl}. 

For data-efficient policy learning, modern simulation platforms such as IsaacSim \cite{mittal2023orbit}, PyBullet \cite{coumans2021pybullet}, Genesis \cite{Genesis}, Robotwin \cite{mu2025robotwin} allow safe parallelized reinforcement learning (RL) exploration \cite{li2017deeprl}. However, sim2real transfer is fundamentally hindered by sparse rewards and exponential exploration complexity in long-horizon tasks \cite{wang2022learning}, and catastrophic error propagation across sequential subtasks owing to compounding inaccuracies. Consequently, data-driven approaches leveraging imitation learning (IL) \cite{hussein2017imitation} and policy distillation (PD) \cite{rusu2015policy} are gaining traction for improved sample efficiency \cite{robomimic2021}, although scaling high-quality teleoperation data remains prohibitively expensive due to human-robot morphological differences \cite{darvish2023teleoperation}. While integrated RL/IL/PD strategies \cite{zhang2025robustdexgrasp,wan2023unidexgrasp++} mitigate data costs, they often overlook the critical influence of diverse object properties and grasp configurations in sim2real deployment.

To overcome these limitations, we introduce a multimodal learning framework for dexterous manipulation. Our primary contributions are threefold: (1) We develop a tactile-based multimodal perception representation capable of maximally leveraging environmental information and proprioception, while dynamically selecting optimal contact force outputs based on object material texture. (2) We construct an asymmetric actor-critic (AAC) network architecture utilizing privileged information; this framework employs simulated privileged data (e.g., deformable object deformation states) for policy value estimation within simulation, enabling optimal control mode selection and reducing the excessive reliance on perceptual precision in contact-intensive operations. (3) We propose a hybrid training paradigm that incorporates multiple category of objects and defines task-specific grasping postures, enabling the agent to acquire enhanced generalization capabilities.

\begin{figure*}[t]
\centering
\includegraphics[width=1.0\textwidth]{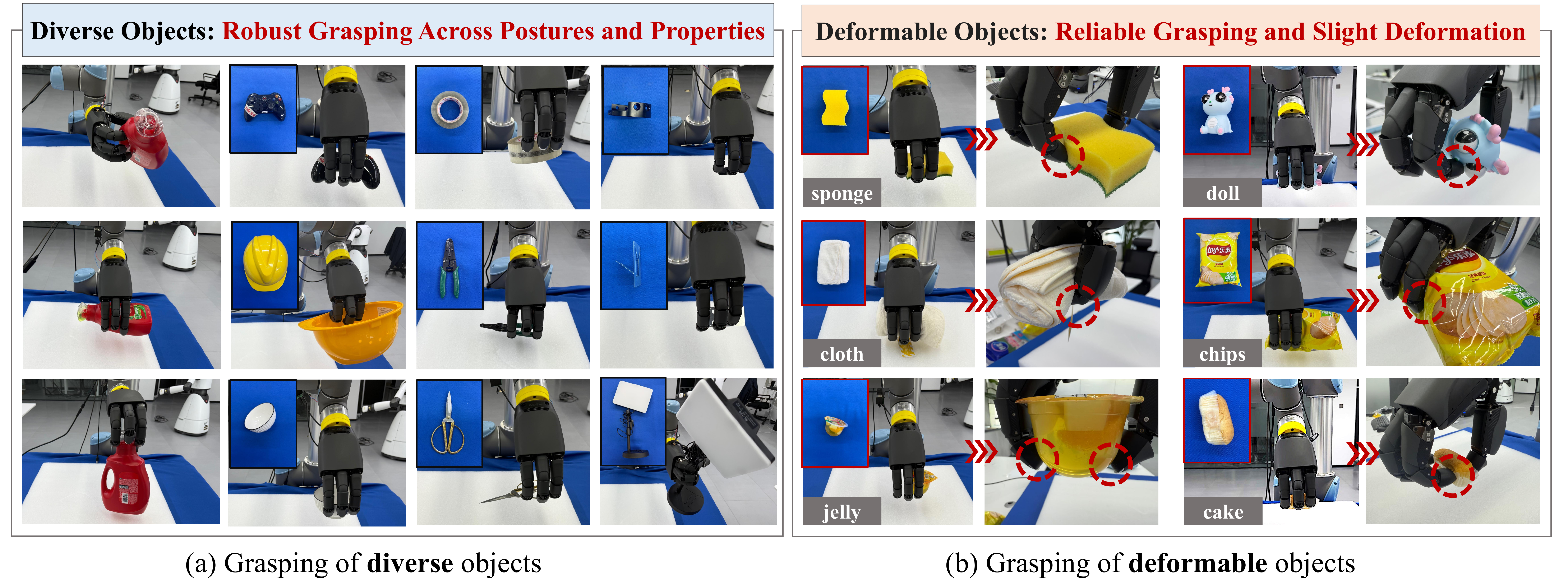} 
\caption{Our method demonstrates robust grasping capabilities for diverse objects, including challenging object categories of diverse sizes, thicknesses, and materials, using different grasping types (a). Furthermore, it effectively minimizes deformation and ensures stable grasps when grasping deformable objects (b). }
\label{fig1}
\end{figure*}

\section{Related Works}

\paragraph{Generative Dexterous Grasping}
Stable grasping with dexterous hands represents a significant challenge in robotics. Early research primarily modeled grasping postures through geometric matching, using feature mapping between hand and object models to establish contact points \cite{wang2023dexgraspnet,shao2020unigrasp}. To enhance stability, subsequent work incorporated traditional hand kinematics to compute force closure at contact points (synthesizing diverse and physically stable grasps with arbitrary hand structures using differentiable force closure estimator), with the aim of resisting external disturbances or enabling complex manipulations. However, these approaches typically rely on explicit mappings between hand and object models, learning grasping posture correspondences from existing datasets. Grasping, being inherently exploratory and dynamic, demands strong generalization and robustness. Although dexterous generative grasping methods \cite{wei2024d, zhang2024contactdexnet, wang2023dexgraspnet, lundell2021ddgc, li2022gendexgrasp} can produce diverse plausible static grasps for various objects, they often struggle to generalize effectively to novel objects and maintain stable anti-interference capabilities under real-world conditions characterized by limited perception and dynamic changes. In contrast, our proposed method enables real-time control of dexterous hand joints primarily through proprioception, facilitating adaptive adjustment of grasp posture, and demonstrating superior robustness.

\paragraph{Learning-based Dexterous Grasping}
In contrast to static grasp generation, recent progress has emphasized data-driven learning methods \cite{li2025maniptrans,yu2023mimictouch,vtguzey2024seetotouch,zhao2024transferable}, which demonstrate promising performance for dexterous grasping of complex objects. However, these methods face limitations stemming from the high cost of real-world dataset acquisition and the challenges in out-of-distribution generalization. Training paradigms such as behavior cloning with expert demonstrations \cite{chen2022dextransfer,chen2024object,li2025maniptrans,li2024okami,fang2025anydexgrasp} frequently suffer from compounding errors due to distributional shift. Alternative approaches include inverse reinforcement learning (IRL) \cite{arora2021irl,orbik2021inverse}, which infers underlying reward functions from demonstrations, and offline reinforcement learning \cite{prudencio2023orl,zhang2023offline}, which learns from static datasets without online interaction. Nevertheless, these techniques require substantial investment for high-quality expert data collection,often involving costly teleoperation, and remain constrained by the quality and diversity of demonstration data. Consequently, RL-based methods \cite{zhang2025robustdexgrasp, choi2023learning,gu2024advancing, lum2024dextrah} leverage the inherent robustness and generalization capabilities of reinforcement learning. Utilizing high-fidelity simulators like IsaacSim \cite{makoviychuk2021isaac} enables cost-effective exploration and real-time adaptation to simulated disturbances. However, this reduced reliance on demonstrations often results in methods that primarily focus on grasping along predefined axes, limiting adaptability to object-specific optimal grasping directions. The persistent sim-to-real transfer challenge is further complicated by high-dimensional state-action spaces, prompting some studies to incorporate human data for training assistance \cite{agarwal2023dexterous,lum2024dextrah,singh2024dextrah} typically requiring known object and hand models during deployment. In contrast, our approach integrates the structured exploration of symbolic reasoning with the adaptability of reinforcement learning and traditional motion planners. This framework addresses limitations of purely data-driven methods while embedding the diversity characteristic of static grasp generation directly into reinforcement learning policies, enabling diverse grasp executions. This synthesis maximizes the rationality of grasping policy under real-world perceptual constraints.

\paragraph{Multimodal Learning}
Within dexterous manipulation, numerous works \cite{vtakinola2024tacsl, matak2022planning} have expanded information perception from single-modality environmental data to richer combinations of proprioceptive and external perception. Only vision-based systems, particularly those leveraging spatial features \cite{higuera2024sparsh,zhao2024transferable,zhang2024dexgraspnet,wang2023dexgraspnet}, excel at global object localization but falter with reflective surfaces and occlusions \cite{vtferrandis2024learning}. Existing research demonstrates that tactile sensing \cite{matak2022planning} can complementarily enhance precision grasping in visually limited scenarios. Early fusion attempts concatenate raw sensor streams, employing self-supervised learning or attention-based feature extractors to encode visual and tactile signals \cite{guzey2023dexterity,vtguzey2024seetotouch,vtdave2024multimodal}. However, these static fusion rules fail to adapt to phase-dependent perceptual correlations, often prioritizing vision for precontact positioning and tactile feedback during insertion \cite{matak2022planning}. Recent neural-symbolic methods \cite{li2022see,wang2025chain,zhang2025embodied} partially address this issue via predefined task templates, yet require manual perceptual mapping for each subtask. Critically, these algorithms typically perform modality selection without correspondingly adapting the core policy learning strategy.

\begin{table*}[t]
\centering
\footnotesize 
\renewcommand{\arraystretch}{1.1} 
\setlength{\tabcolsep}{2pt} 
\newcommand{\greencheck}{{\color{Green}$\checkmark$}}
\begin{tabular}{l|c|c|c|c|c|c}
 \hline
 \Xhline{0.5pt}
\textbf{Method} & \textbf{Taxonomy} & \textbf{MultiModal} & \textbf{Sim2Real} & \textbf{Dexterous} & \textbf{Deformable} & \textbf{Diverse} \\ \hline
 \Xhline{0.5pt}
DexGraspNet2.0 \cite{zhang2024dexgraspnet}  & Gen & V & $\checkmark$ & LEAP hand (0 Sensors) & $\times$ & random \\
D(R,O) \cite{wei2024d}& Gen & V & $\checkmark$ & Allegro+Shadowhand (5 Sensors) & $\times$ & random \\
D-Grasp \cite{christen2022d}& RL & P & $\times$ & Simulated hand (0 Sensors) & $\times$ & 3 \\
UniDexGrasp++ \cite{wan2023unidexgrasp++}  & RL+IL & V+P & $\times$ & Simulated hand (0 Sensors) & $\times$ & 1 \\
DextrAH-G \cite{lum2024dextrah} & RL & V+P & $\checkmark$ & Allegro Hand (4 Sensors) & $\times$ & 1 \\
RobustDexGrasp \cite{zhang2025robustdexgrasp} & RL & V+P+T* & $\checkmark$ & Allegro Hand (4 Sensors) & $\times$ & 1 \\
\textbf{D3Grasp (Ours)} & RL & V+P+T & $\checkmark$ & Dexhand13 (11 Sensors) & $ \checkmark$ & 3 \\
 \hline
\Xhline{0.5pt}
\end{tabular}
\caption{Comparison with contemporary dexterous grasping approaches. \textbf{Taxonomy}: generative methods (Gen), reinforcement learning (RL) and imitation learning (IL). \textbf{MultiModal}: V is visual information (point clouds or images), P is proprioception (joint angles), T is tactile perception, where T* is synthesized tactile signals. Notably, our framework utilizes solely visual inputs for object positioning yet maintains blind grasping capability in perceptually limited environments.\textbf{Dexterous}: Dexterous hands with the number of tactile sensors include Allegro \cite{allegrohand}, Barrett \cite{Barrett}, Shadow Hand \cite{shadowhand2013}, LEAP Hand \cite{shaw2023leaphand}, and DexHand-13 \cite{paxinidexhand}, representing current mainstream dexterous manipulators . \textbf{Diverse}: the directional diversity of grasping strategies generated by the method.} 
\label{table1}
\end{table*}
\section{Methods}
\begin{figure*}[t]
\centering
\includegraphics[width=1.0\textwidth]{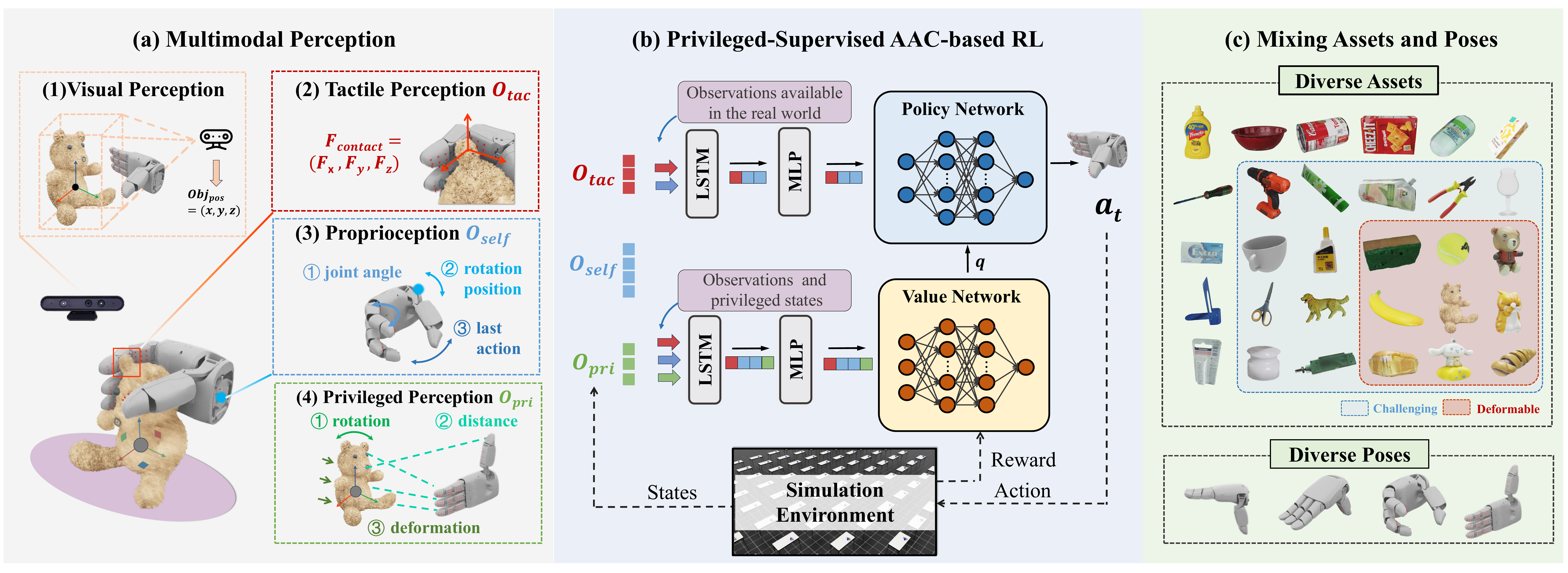} 
\caption{Framework overview. (a) We introduce a multimodal perceptual representation. (b) We independently acquire multimodal perceptual information in a simulated environment and encode it through a feature extraction encoder, using the embedding as the input of AAC network. The Actor network predicts subsequent continuous 16-DoF joint of commands $a_t$ for the hand. (c) During the training, our approach involves replacing training environment assets and poses, incorporating deformation feedback specific to deformable objects, and enhancing the agent's generalization capabilities.}
\label{fig:model}
\end{figure*}

\label{sec:citations}
We aim to leverage multimodal perception incorporating tactile sensation to guide a dexterous manipulation system within perceptually constrained scenarios. The proposed multimodal guided dexterous operating system is illustrated in Fig.\ref{fig:model}
we constructed a dedicated simulation environment for the training of dexterous hand policies, featuring high-fidelity modeling of the geometric structures, joint dynamics, and contact force sensor arrays of the dexterous hand.

\subsection{Multimodal Perception}

Multimodal perception is critical for enabling adaptive real-time grasping during dynamic interactions, particularly in perceptually constrained unknown environments. Mimicking the cognitive hierarchy of human grasping behavior, the visual modality first provides coarse-grained spatial localization. Utilizing an RGB-D sensor, this generates the 3D workspace coordinates and approximate geometric contours of the target object,which enables the determination of the end-effector pose through solving inverse kinematics (IK).

In the subsequent phase, we define the observational state representation for Reinforcement Learning (RL) training, serving as the core sensory source for grasping. First, we employ 11 tactile sensors on finger joints to capture three-dimensional contact forces, enabling real-time monitoring of fine-grained local interactions during grasping. Second, we incorporate egocentric motion perception to guide the manipulation process; this self-perception modality includes the dexterous hand's current joint positions, linear/angular velocities, and action histories to model action-effect relationships, thereby ensuring behavioral continuity. 

Finally, privileged perception within simulation provides access to real-world unavailable/costly information (e.g., object motion states, deformable object strain fields, and hand-object distances). The simulation of deformable objects in Isaac Sim employs the Finite Element Method (FEM) to discretize objects into grids comprising nodes and elements. Consequently, representing deformable objects in the simulation environment requires the computation of nodal velocities and displacements. For an accurate estimation of an object's global rotation, we implement an adaptive rotation computation method based on the object's deformation degree. First, we compute the standard deviation $\sigma$ of the Euclidean distance between the position of each node $\mathbf{p}_i$ and its reference position $\mathbf{p}_i^{\text{ref}}$ as a deformation metric:
$\sigma = \operatorname{Std}( \{ \| \mathbf{p}_i - \mathbf{p}_i^{\text{ref}} \|^2 \}_{i=1}^N)$.
The rotation calculation strategy is then determined as follows.
\begin{equation}
q_{\text{do}} =
\begin{cases}
\text{Average\_rotation} & \sigma < \tau_{1} \\
\text{Principal\_rotation}  & \tau_1 \leq \sigma < \tau_2 \\
\text{Rigid\_rotation} & \sigma \geq \tau_2
\end{cases}
\end{equation}

\noindent where $\tau_1 = 0.01$ and $\tau_2 = 0.1$ represent deformation thresholds (implementation details are given in the Appendix). For linear velocity computation, we derive the object's global linear velocity from its center-of-mass (CoM) motion $\mathbf{s}_c$:
\begin{equation}
    \mathbf{v}_{\text{do}} = [{ \mathbf{s}_c(t+\Delta t) - \mathbf{s}_c(t) }]/{ \Delta t }
\end{equation}

For object deformation quantification, we compute the Green-Lagrange strain tensor $\mathbf{E}$ from simulated deformation gradients $\mathbf{F}$, subsequently deriving the maximum principal strain (dominant eigenvalue of $\mathbf{E}$ at nodal evaluations) to characterize the peak tensile deformation at material points.
\begin{equation}
\begin{aligned}
& \mathbf{E} = \frac{1}{2} (\mathbf{F}^{\!\top} \mathbf{F} - \mathbf{I}) \\
& \varepsilon_{\max} = \max \{\lambda_i \mid i=1,2,\dots,n\}\\
& \bar{\varepsilon}_{max} = \frac{1}{N_k} \sum_{i}^{N_k} \varepsilon_{\max}^{(i)}    
\end{aligned}
\end{equation}

\noindent where $\lambda_i$ is the eigenvalue of the strain tensor E, $N_k$ is the number of simulated nodes in deformable objects.

\subsection{Privileged-Supervised AAC}

Our approach primarily employs asymmetric reinforcement learning, utilizing Critic networks that access the full perception including privileged information available in the simulation environment. Subsequently, we train our Actor network on realistic dexterous hand perception.

\noindent\textbf{Observation.} At each control cycle $i$, the system collects multimodal data, forming the observation vector$s_i = [o_{tac},o_{self},o_{pri}, t_i]$. Here, $o_{self}$ denotes proprioceptive state; $o_{tac}$ represents tactile perception, $o_{pri}$ encapsulates privileged information, and $t_i$ encodes both the current time step and its corresponding sinusoidal (sine and cosine) embedding to provide explicit temporal awareness.
To model the inherent sequential dependencies and temporal correlations in dexterous manipulation, the observation vector $s_i$ is first processed through a Long Short-Term Memory (LSTM) network. The resulting hidden states are subsequently fed into a Multilayer Perceptron (MLP), which extracts high-level spatiotemporal features suitable for effective policy learning. These features are provided as input to the Proximal Policy Optimization (PPO) agent.

\noindent\textbf{Reward.} To constrain general grasping behavior and promote the accomplishment of specific grasping tasks, a comprehensive reward function is designed, comprising components for generality, distance, appropriate contact, and stability:
\begin{equation}
\begin{aligned}
 reward &= r_\text{generic} + r_{\text{contact}} + r_{\text{distance}} \\& +  r_{\text{symmetry}} + r_{\text{stability}} \\ 
 & + r_{\text{collision}}
\end{aligned}
\end{equation}

The general reward $r_{\text{generic}}$ promotes safe and energy-efficient motion by rewarding system integrity while penalizing excessive displacement and velocity of the joints:

\begin{equation}
   r_{\text{generic}} = k_{a}f_\text{alive} + k_{j}||j||^2+k_{jv}||v_{j}||^2
\end{equation}

\noindent where $f_\text{alive}$ indicates the boolean survival status and $j$,$v_{j}$ are joint displacements and motion velocities. The distance reward $r_\text{distance}$ incentivizes agents to approach the object:

\begin{equation}
    r_{\text{distance}} = -\sum_{i}^{N} k_{i} ||{d}_i ||^2
\end{equation}

\noindent where ${d}_i$ denotes the Euclidean distance between the $i$-th hand joint and the object's center of mass, and $N$ is the joint number. The contact reward encourages sufficient and reasonable contact between the fingertips and the object surface while constraining the magnitude of the contact force.
\begin{equation}
    r_{\text{contact}} = \sum_{i}^{N} [b_ik_{fi} f_i - k_{pi} (f_i - f_m)  ]
\end{equation}
\noindent Here, $b_i$ is a binary indicator of contact force exceeding a minimum threshold in the joint $i$, $f_j$ is the contact force, $f_m$ is the maximum permitted force, and $k_{fj}$, $k_{pi}$ are weighting coefficients for contact establishment, force magnitude and penalty for violation of force. Respectively,stability reward penalizes extra object motion and minimizes physical deformation to maintain structural integrity of the object.

\begin{equation}
\begin{aligned}
    &r_{\text{stability}}= -k_{ov} || {v}_o ||^2 - k_{o\omega} ||{\omega}_o||^2
\end{aligned}
\end{equation}

\noindent where ${v}_o$ , ${\omega}_o$ are the linear and angular velocities of the grasped object. Finally, undesired environmental interactions are discouraged by collision reward, $d_{ie}$ is the distance from joint $j$ to the environment and $m_{ie}$ represents a predefined minimum safe distance threshold.   

\begin{equation}
    r_{\text{collision}} = -\sum_{i}^{N} \log(\min\{d_{ie}, m_{ie}\})
\end{equation}

\subsection{Diverse Training with Mixing Assets and Poses}
To improve generalization in complex environments, the training process incorporates diverse assets with mixed grasp poses, as illustrated in Figure \ref{fig:model} (c). Crucially, we incorporated deformable objects with varying elastic moduli and material textures into the training assets, introducing $r_\text{deform}$ to restrict excessive deformation during training.
\begin{equation}
r_{\mathrm{deform}} =  
\begin{cases} 
    -k_d \epsilon_o = -k_d N_k \bar{\varepsilon}_{\max}, & \text{deformable object} \\
    0, & \text{rigid object}
\end{cases}
\end{equation}

Given the need for task-specific adaptations to grasping postures based on object geometry and functional requirements, dexterous manipulation requires targeted posture optimization. The high-dimensional action space inherent to reinforcement learning frameworks can impede convergence or cause policy instability. Consequently, our approach focuses on selecting some archetypal grasp categories and identifying biomechanically feasible postures for distinct object types during training. This methodology reduces implausible grasp executions in real-world scenarios while enhancing training efficiency.

\section{Experiments}


\subsection{Experimental Setup}

For the grasping part, we employ the PPO algorithm based on RL\_games \cite{makoviichuk2021rl} to train the strategy. The learned grasping policy is then transferred to a physical dexterous hand for real-world evaluation (more details in Sec \ref{eval}). We import additional assets from other benchmarks \cite{asset,makoviychuk2021isaac} to augment our simulation environment, leveraging a diverse set of more than \textbf{100} object instances for comprehensive model training and evaluation. Furthermore, our method is evaluated on more than \textbf{50} distinct physical objects in real-world validation.

\begin{figure}[tbh]
\centering
\includegraphics[width=0.6\columnwidth]{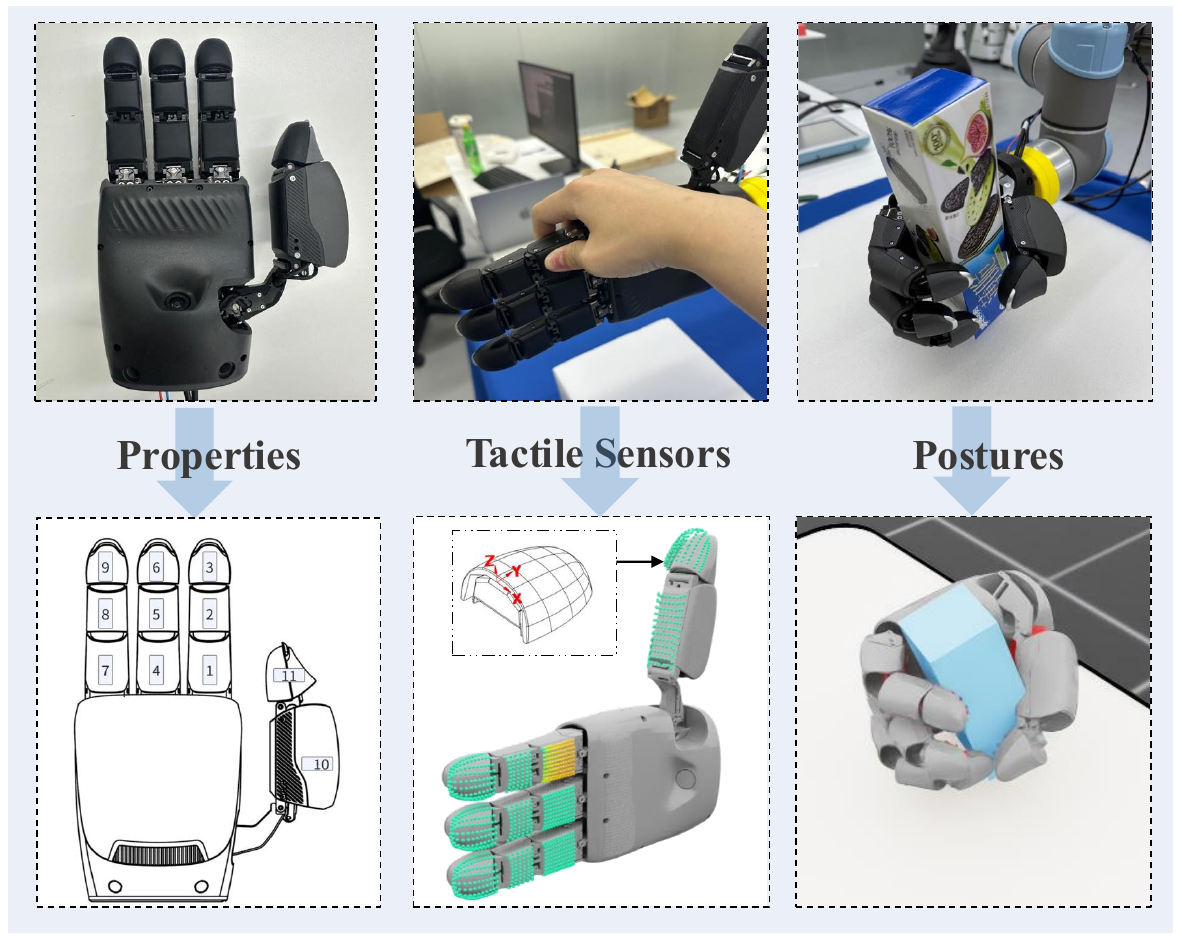} 
\caption{Simulation configs. Instructions for the tactile sensor and joint settings of Dexhand13 with grasping instantiation in simulation. }
\label{fig:simulation}
\end{figure}

\noindent\textbf{Simulation.} The simulation is built using IsaacSim \cite{makoviychuk2021isaac} for a dexterous hand simulation. The hand model is based on the Paxini DexHand13 \cite{paxinidexhand}, which has four fingers and a total of 16 DoF. The root pose also guides the 6D pose of the UR5 arm, resulting in a total of 22 DoF. In simulation, we only modify the 6D root pose of the dexterous hand, while in the real world these actions directly guide the movements of the robotic arm.Regarding sensing, the hand is equipped with 11 tactile sensors, as illustrated in Fig. \ref{fig:simulation} . These tactile sensors provide accurate 3D force matrices with a defined coordinate system at the origin of the pad’s base.

\begin{figure}[t]
\centering
\includegraphics[width=0.7\columnwidth]{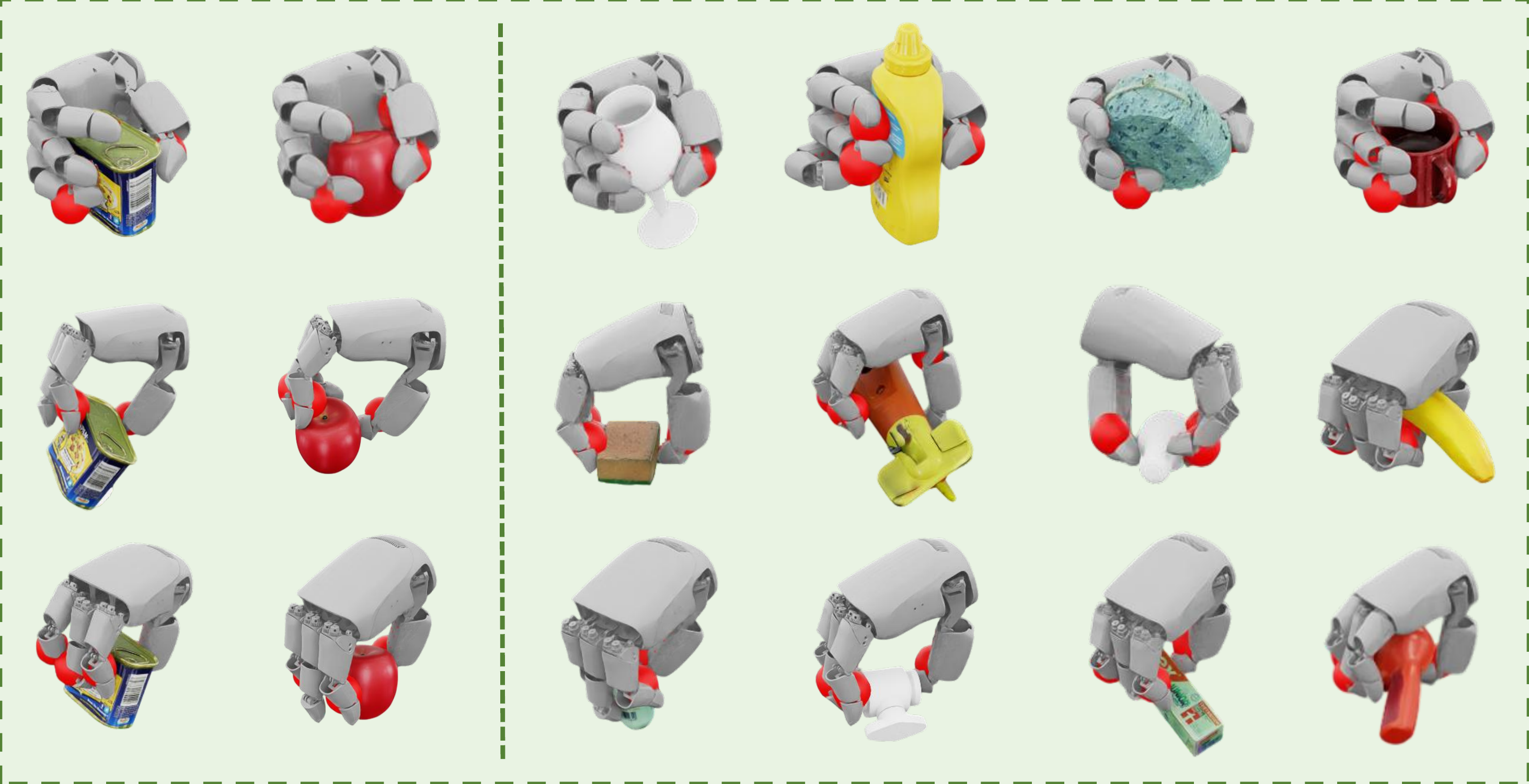}
\caption{Dexterous hand robustly grasps diverse objects in the sim environment.}
\label{fig：simresult}
\end{figure}

\noindent\textbf{Task.} We focus on the reach-grasp-lift  task, which involves three critical subtasks: reaching for an object, grasping, lifting and release it. This task encapsulates the core challenges of dexterous manipulation(e.g., precise arm positioning, grasp stability, and adaptive force control), which are essential for real-world applications like object retrieval and assistive tasks. Furthermore, this task can be easily generalized to a wide range of manipulation scenarios, making it a good benchmark to test the performance of dexterous manipulation systems. 

\noindent\textbf{Metrics.} To ensure robust anti-interference capabilities while grasping objects, we use three evaluation metrics: (1) Grasp success rate $s_{\text{grasp}}$, determined by tactile joint feedback indicating proper hand-object contact; (2) Lift success rate $s_{\text{lift}}$, recorded when objects maintain stability lifted to the height of 0.1m ; (3) Disturbance resistance success rate $s_{\text{disturb}}$, measured by applying 2.5N external forces along six axes for 2 s after lifting and the object moves below 0.02 m.

\subsection{Compared with other methods}
\begin{table*}[htb]
\centering
\footnotesize 
\renewcommand{\arraystretch}{1.1} 
\setlength{\tabcolsep}{1pt} 
\begin{tabular}{c|cccc|c|cccc}
\hline
\Xhline{0.5pt}
\multirow{3}{*}{Method} & \multicolumn{4}{c|}{Simulation} & \multirow{3}{*}{Method} & \multicolumn{4}{c}{Real World} \\ 
\cline{2-5} \cline{7-10}
& \multicolumn{2}{c|}{Diverse} & \multicolumn{2}{c|}{Deformable} & &\multirow{2}{*}{Obj.(num)} & \multirow{2}{*}{$s_{\text{lift}}$} & \multirow{2}{*}{$s_{\text{disturb}}$} & \multirow{2}{*}{Eff.(sec)} \\  
\cline{2-5} 
& $s_{\text{lift}}$ & \multicolumn{1}{c|}{$s_{\text{disturb}}$} & $s_{\text{lift}}$ & $s_{\text{disturb}}$ & &  \multicolumn{4}{c}{} \\  
\hline
Rule-based & 55.0 & 41.4 & 48.3 & 39.5 & Rule-based & 50 & 56.1 & 52.2 & - \\
DexGraspNet \cite{wang2023dexgraspnet} & 66.7 & 55.3 & - & - & DexGraspNet \cite{wang2023dexgraspnet} & 30 & 60.7 & 48.0 & - \\
DexDiffuser \cite{weng2024dexdiffuser} & 64.0 & - & 60.0 & - & SpringGrasp \cite{chen2024springgrasp}& 30 & 77.1 & - & - \\
ISAGrasp \cite{isagrasp} & 63.4 & - & 60.0 & - & $\mathcal{D(R,O)}$ \cite{wei2024d}& 10 & 89.0 & - & 0.65 \\
D3Grasp (Ours) & \textbf{84.2} & \textbf{82.7} & \textbf{84.4} & \textbf{82.2} & D3Grasp (Ours) & \textbf{50} & \textbf{91.0} & \textbf{90.5} & \textbf{0.36} \\
\hline
\Xhline{0.5pt}
\end{tabular}
\caption{Comparison results with other methods.We quantify \textbf{Eff.}(Efficiency) by measuring the latency of our method's 200-step reasoning process as the key performance metric. \textbf{Obj.} refers to the number of assets in real-world evaluation.}
\label{table:compared}
\end{table*}
To evaluate the robust generalizability and resilience of our method in multimodal environments, we conducted comparisons with established baselines within the same simulated setting. In particular, real-world experimental results for these baseline methods were previously reported in related work \cite{zhang2025robustdexgrasp,wang2023dexgraspnet}: (1) rule-based method: hand joints are closed at a constant speed,
and when the tactile sensors on the joints exceed the con-
tact force threshold, the grasping is considered successful. (2) DexGraspNet \cite{wang2023dexgraspnet}: A grip generation
method based on the Isaac Gym simulation
environment. \cite{makoviychuk2021isaac} (3) SpringGrasp \cite{chen2024springgrasp} :
a optimization-based planner that generates dexterous grasps considering the object's obeservation. (4) $\mathcal{D(R,O)}$ \cite{wei2024d}: a framework that models the interaction between the robotic hand in its grasping pose and the object. (5) DexDiffuser \cite{weng2024dexdiffuser}: a dexterous
grasping method that generates grasps on partial object point clouds with diffusion models. (6) ISAGrasp \cite{isagrasp}: a generative model that deforms object meshes and human grasps. The results are shown in Table. \ref{table:compared}

\subsection{Evaluation}
\begin{figure*}[tbh]
\centering
\includegraphics[width=1.0\textwidth]{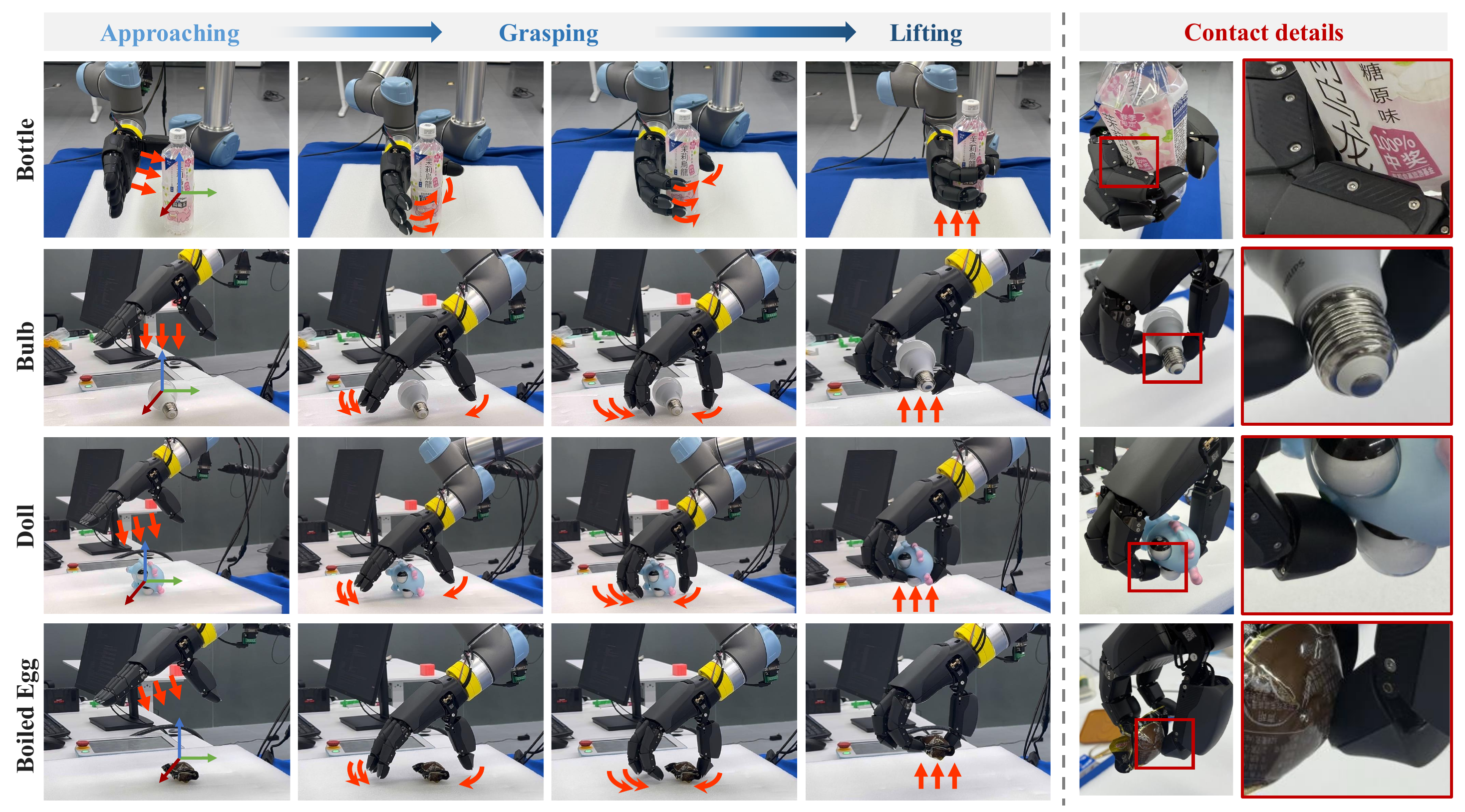} 
\caption{This visualization illustrates the complete process of our method from the approach phase to lifting, with magnified details of the grasping contact during the lifting process. }
\label{fig:visual}
\end{figure*}

\label{eval}
\begin{table*}[htb]
\centering
\renewcommand{\arraystretch}{1.1} 
\setlength{\tabcolsep}{3pt} 
\begin{tabular}{c|c|c|c|c|c|c|c|c|c|c}
\hline
\Xhline{0.5pt}
    \multirow{3}{*}{Metric(\%)} & \multicolumn{5}{c|}{Simulation} & \multicolumn{5}{c}{Real World} \\ 
      \cline{2-11}
      & \multicolumn{3}{c|}{Diverse Obj.} & \multirow{2}{*}{Deformable Obj.}&  \multirow{2}{*}{Avg.}& \multicolumn{3}{c|}{Diverse Obj.} & \multirow{2}{*}{Deformable Obj.}&  \multirow{2}{*}{Avg.}\\ 
      \cline{2-4} \cline{7-9}  
      &Small &  Medium    & Large & & &Small &  Medium    & Large & & \\ \hline 
      $s_{\text{grasp}} $ & 81.5 & 90.2&92.7&87.8 & 87.9 & 90.1 & 97.3 & 98.3 & 95.0 & 95.1 \\
      $s_{\text{lift}} $ & 72.4 & 89.0 & 91.4& 84.4 & 84.3 & 85.1 & 93.9 & 90.0 &93.1& 91.0 \\
      $s_{\text{disturb}}$ & 70.7 & 87.6& 89.5 & 82.2 &82.6 & 82.2 & 92.6 & 83.3 & 90.0  & 90.5 \\
\hline
\Xhline{0.5pt}
\end{tabular}
\caption{Evaluation results in simulation and real world environments.}
\label{table:evaluation}
\end{table*}

\textbf{Simulation.} As for simulation , we focus solely on learning the grasping, where the root position of the dexterous hand and the target object are initialized and fixed. Fig. \ref{fig:simulation} shows some of the grasping results in the simulation and
we performed evaluations utilizing various 3D assets in IsaacSim \cite{makoviychuk2021isaac}. Given the inherent difficulty in uniformly determining the optimal lifting timing for diverse object types, our approach trains the agent to autonomously maintain a stable grasping posture conducive to successful lift. Consequently, grasping success is primarily determined by the sustained holding duration. Simultaneously, to mitigate abnormal hand-object interactions, we employ elementary tactile perception to enhance lifting timing determination. The evaluation results are presented in the Table. \ref{table:evaluation}.

\noindent\textbf{Real-world.} For sim2real transfer evaluation. We have developed a sim2real system using the UR5 \cite{ur5} robotic arm equipped with the Dexhand13 dexterous hand to perform grasp-lift experiments in the real world.
During the lift phase, we introduce controlled external disturbances to assess grasping. Successful execution requires completion of both the grasp and the lift phases within fixed steps $n$ . Failure to satisfy this condition terminates the trial. To accommodate heightened contact friction in physical dexterous hands, we reduced contact force thresholds during real-world validation. This adjustment enables rigorous assessment of the hand's ability to achieve static equilibrium and successful lifting under placement inaccuracies and observation noise, while demonstrating robust object transfer to target locations without slippage or damage.

We conducted ten grasping trials per object, with experimental results detailed in Table. \ref{table:evaluation}. Our system demonstrated robust performance through real-world evaluation. The RL agent exhibited strong generalization capabilities across both rigid and deformable objects through adaptive grasping strategies. Utilizing the high-sensitivity tactile feedback of the Dexhand13 manipulator, the system infers optimal joint configurations during grasping operations, thereby preventing damage to deformable objects (e.g., sponges and chips) while maintaining stable grasps.
\subsection{Ablation Study}

\begin{table}[htb]
\renewcommand{\arraystretch}{1.25}
\setlength{\tabcolsep}{9mm}
\begin{tabular}{c|cc|c}
 \hline
 \Xhline{0.5pt}
\multirow{3}{*}{Metric(\%)} & 
\multicolumn{2}{c|}{Symmetric} & 
\multicolumn{1}{c}{Asymmetric} \\
 \cline{2-4}
 & \multicolumn{1}{c}{\multirow{2}{*}{Base}} & 
 \multicolumn{1}{c|}{\multirow{2}{*}{Base+Tactile}} & 
 \multicolumn{1}{c}{Base+Tactile (Actor)} \\
 & &  & 
 \multicolumn{1}{c}{\multirow{1}{*}{Base+Tactile+Privileged(Critic)}} \\
 \hline
 $s_{\text{grasp}}$ & 87.2 & 85.1  & \textbf{95.1} \\
 $s_{\text{lift}}$ & 56.4 & 65.7 & \textbf{91.0} \\
 $s_{\text{disturb}}$ & 49.8 & 63.2  & \textbf{90.5} \\
 \hline
 \Xhline{0.5pt}
\end{tabular}
\caption{Ablation analysis of grasping in real world.  }
\label{table:ablation}
\end{table}
We perform a comparative evaluation of grasping performance between symmetric and asymmetric network architectures under equivalent training conditions, while systematically assessing the impact of the incorporation of tactile perception. Quantitative results are presented in Table. \ref{table:ablation}.

Analysis of various input modalities and network architectures reveals that relying solely on proprioceptive information results in the failure to learn effective grasping strategies. However, the integration of 3D tactile perception significantly enhances the grasping success rate. This underscores the critical importance of tactile feedback, particularly in contact-intensive tasks where shape deformation and friction substantially influence grasping outcomes. Furthermore, the AAC network with privileged input exhibits a higher convergence rate and superior grasping stability.
\section{Conclusion and Future Work}
In this work, we present D3Grasp, a multimodal perception-guided framework for dexterous manipulation that enables diverse and adaptive grasp strategies across a wide range of objects, with particular robustness for deformable and contact-sensitive cases. Leveraging cross-modal visual-tactile perception, asymmetric actor-critic based reinforcement learning, and progressive curriculum learning, our approach effectively balances grasp diversity, stability, and adaptability, particularly for deformable and contact-sensitive objects. Without relying on manual demonstrations or complex object modeling, D3Grasp learns directly from simulated interactions while maintaining strong sim-to-real generalization. Extensive experiments across simulation and real-world deployments validate the system's ability to robustly execute complex grasping behaviors under perceptual constraints. These results position D3Grasp as a highly robust, deployment-friendly and general-purpose solution for practical dexterous robotic manipulation.

\bibliographystyle{assets/plainnat}
\bibliography{paper}



\end{document}